\documentclass[12pt]{article}

\usepackage{macros}

\usepackage{xcolor}

\usepackage{amsopn}

\usepackage{lmodern}
\usepackage{amsmath}

\usepackage{graphicx}

\usepackage{amssymb,amsmath}

\usepackage{relsize}

\usepackage{upgreek}

\usepackage{subfigure}

\usepackage{url}

\usepackage{graphicx}

\usepackage{algorithmic}
\usepackage{algorithm2e}

\pdfoutput=1
\usepackage{amsmath, amssymb}
\usepackage{graphicx,psfrag,epsf}
\usepackage{enumerate}
\usepackage{natbib}
\usepackage{url} 

\def\x{\mathbf{x}}

\def\R{\mathbb{R}}

\def\Z{\mathbf{Z}}
\def\X{\mathbf{X}}

\def\V{\mathbf{V}}

\def\z{\mathbf{z}}
\def\nb{Na{\"i}ve Bayes}
\def\pp{{p^\prime}}

\def\one{\mathbf{1}}

\begin{document}

\def\spacingset#1{\renewcommand{\baselinestretch}%
{#1}\small\normalsize} \spacingset{1}


  \title{\bf Optimal Projections for Classification with Na{\"i}ve Bayes}
  \author{David P. Hofmeyr
  \hspace{.2cm}\\
    School of Mathematical Sciences, Lancaster University, United Kingdom\\
    Francois Kamper\\
    Swiss Data Science Centre, EPFL \& ETH Z{\"u}rich, Switzerland\\
    Michail C. Melonas\\
    Kohort, Cape Town, South Africa}
  \maketitle

\bigskip
\begin{abstract}
In the \nb~classification model the class conditional densities are estimated as the products of their marginal densities along the cardinal basis directions. We study the problem of obtaining an alternative basis for this factorisation with the objective of enhancing the discriminatory power of the associated classification model. We formulate the problem as a projection pursuit to find the optimal linear projection on which to perform classification. Optimality is determined based on the multinomial likelihood within which probabilities are estimated using the \nb~factorisation of the projected data. Projection pursuit offers the added benefits of dimension reduction and visualisation. We discuss an intuitive connection with class conditional independent components analysis, and show how this is realised visually in practical applications. The performance of the resulting classification models is investigated using a large collection of (162) publicly available benchmark data sets and in comparison with relevant alternatives. We find that the proposed approach substantially outperforms other popular probabilistic discriminant analysis models and is highly competitive with Support Vector Machines.

Code to implement the proposed approach, in the form of an {\tt R} package, is available from \url{https://github.com/DavidHofmeyr/OPNB}.
\end{abstract}

\spacingset{1.45} 

\section{Introduction} \label{sec:intro}

Suppose we are presented with pairs $(y_1, \x_1), ..., (y_n, \x_n)$ assumed to have arisen independently from some joint probability distribution, $P_{Y, X}$, on $[K] \times \R^p$, where we have used $[K]$ to denote the first $K$ natural numbers, i.e., $[K] = \{1, ..., K\}$. That is, the {\em class labels}, $\{y_1, ..., y_n\}$, each takes one of $K$ known and distinct values and the associated vectors of {\em covariates}, $\{\x_1, ..., \x_n\}$, are each $p$-dimensional real valued vectors. The problem of discriminant analysis in a probabilistic framework is to obtain an estimate of the {\em posterior probability} of class membership, $P(Y = k | X = \x)$, for each $k \in [K]$, based on a simple application of Bayes' rule,
\begin{align}\label{eq:bayes}
    P(Y = k |X = \x) &= \frac{\pi_k f_{X|Y=k}(\x)}{\sum_{j=1}^K \pi_j f_{X|Y=j}(\x)},
\end{align}
where $\pi_k = P(Y = k)$ is the {\em prior probability} for class $k$, and we use the general notation ``$f_Z$'' to represent the probability density function\footnote{For simplicity we discuss the scenario in which $X$ is a continuous random variable, but acknowledge the scope for a more general formulation of the posterior class probabilities in the presence of discrete or mixed covariates. Note also that because our approach is based on the marginal distributions of (non-sparse) linear projections of $X$, the conditions under which a formulation in terms of continuous $X$ is appropriate are quite general. We provide a brief discussion of this in Section~\ref{sec:OHE}.} of the random variable $Z$. Different approaches to the problem vary according to how they estimate the densities $f_{X|Y = k}$, where it is almost universal that the prior probabilities are estimated using $\hat \pi_k = \frac{n_k}{n}$, for $n_k = \sum_{i=1}^n \one(y_i = k)$, where $\one(\cdot)$ is the indicator function. Popular approaches include treating each such {\em class conditional density} as a Gaussian density with appropriately estimated mean vector and covariance matrix~\citep{citeLDA, citeQDA}, and using non-parametric density estimators~\citep{hand1982}.

The non-parametric variant of discriminant analysis is appealing for its flexibility. However, this flexibility comes at the cost of increased estimation variance, as well as computational complexity. A popular simplification, referred to as {\em \nb}, addresses some of these limitations of density estimation by treating the densities of the random variables $X|Y=k; k \in [K]$, as admitting a simple factorisation over their margins. That is, $\hat f_{X|Y=k} = \prod_{d=1}^p \hat f_{X_d|Y=k}$, where $X_d$ is the $d$-th component of the $p$-dimensional $X$, and $\hat f_{X_d|Y=k}$ is a univariate density estimator. This {\em class conditional independence} approach generally introduces bias into the estimated densities, however it has been observed that in many cases the effect which this has on the accurate estimation of the class decision boundaries is fairly minimal. The class decision boundaries are the surfaces of the sets
\begin{align}
    \left\{\x \in \R^p \bigg| \mathrm{argmax}_{k\in [K]} P(Y = k|X = \x) = j\right\}; j \in [K].
\end{align}
That is to say, from the point of view of accurate classification, our primary concern is not the most accurate estimation of the probabilities $P(Y = k|X = \x); k \in [K]$, but rather which of these probabilities is dominant, i.e., of $\mathrm{argmax}_{k \in [K]} P(Y = k|X = \x)$; and in many instances this estimation problem suffers less from the \nb~formulation than does the class density estimation problem itself.

In this paper we introduce a novel approach for enhancing the performance of \nb~classifiers, in which, rather than applying the standard \nb~factorisation of the class densities over the cardinal basis (i.e., over the coordinate dimensions of $\R^p$), we seek to find an optimal basis over which to perform this factorisation. In particular, we formulate the problem as a projection pursuit in which the objective is given by the multinomial likelihood, where probabilities in this likelihood are determined according to Eq.~(\ref{eq:bayes}), with class densities estimated using the \nb~factorisation on the projected data. This approach has some similarities with the approach of \cite[CCICA]{CCICA}, who apply Independent Components Analysis~\citep[ICA]{comon_ica} to the observations from each of the classes separately before applying a \nb-like factorisation of the class density along its ICA basis directions. 
However, our approach has some very important advantages: (i) since the basis for factorisation is optimised for the classification objective, the predictive ability of the proposed method is vastly superior in general, as we illustrate by means of a large set of experiments in Section~\ref{sec:experiments}; (ii) since there is a single basis (and projection), rather than a separate basis for each class, and the basis is optimised for classification, our model is more interpretable; and (iii) in a related point our model offers pleasing and instructive visualisations of class separability/inseparability, making any failures of the model far easier to diagnose.

\begin{figure*}
    \centering
    \includegraphics[width = \textwidth, height = .22\textwidth]{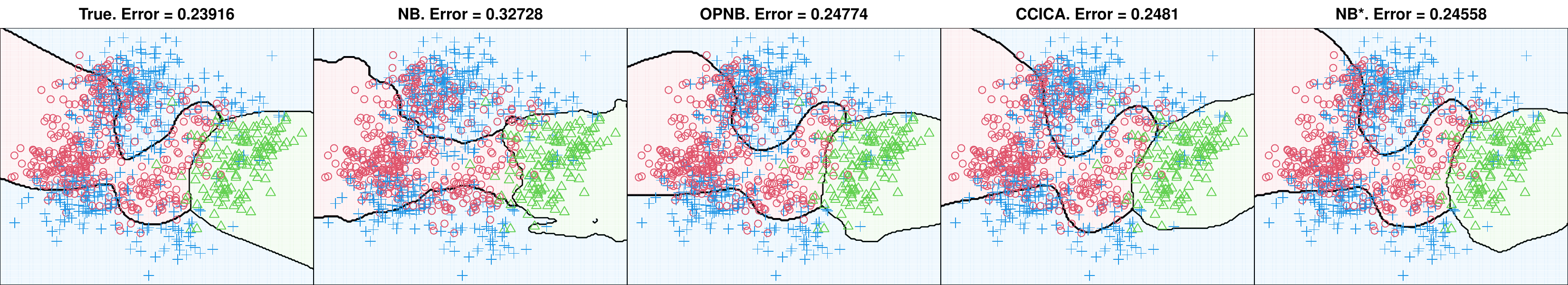}
    \caption{Two-dimensional example in which the class conditional densities factorise over the cardinal basis rotated by $\pi/8$ radians. From left the true class decision boundaries; those estimated using \nb~(NB) as well as the proposed approach (OPNB); Class Conditional Independent Components Analaysis (CCICA); and the \nb~factorisation applied on the cardinal basis rotated by $\pi/8$ (i.e., the optimal rotation at a population level) are shown. The errors correspond with estimates of the test classification error from samples of size 50000.}
    \label{fig:2D}
\end{figure*}

To illustrate the main idea underlying our approach, a simple two-dimensional example is shown in Figure~\ref{fig:2D}. The figure shows a scenario with three classes in which the class conditional densities factorise along the cardinal basis rotated by $\pi/8$ radians. The plots show the same sample of size 1000 drawn from the underlying distribution, with points from different classes differentiated according to colour and point character. The plots also show the true class decision boundaries as well as those estimated from the sample using \nb~(NB); the proposed approach (Optimal Projections for \nb, OPNB); the CCICA model; and the \nb~factorisation applied to the ``true'' ($\pi/8$ rotated cardinal) basis (NB*). Even this relatively moderate rotation results in poor performance by the standard \nb~model, whereas OPNB and CCICA obtain decision  boundaries close to those of NB*, and all three have classification error close to the Bayes optimal error rate of 0.23916\footnote{The classification error of the different models was estimated from a test sample of size 50000, and so may differ very slightly from the population level error rates.}.


The remaining paper is organised as follows. In Section~\ref{sec:methodology} we provide an explicit description of our problem, and provide technical details of the associated optimisation. We also provide further discussion on the connection between OPNB and CCICA~\citep{CCICA}. In Section~\ref{sec:experiments} we discuss results from an extensive set of experiments with the proposed method. We also document the visualisation capabilities which are offered as a result of finding an optimal projection/basis for the data, and how this can aid in moidyfing the model in order to improve performance. We conclude the paper with a discussion in Section~\ref{sec:conclusions}.

\section{Optimal Projections for Discrimination with Na{\"i}ve Bayes Probabilities} \label{sec:methodology}

In this section we provide an explicit formulation of the objective we address, and discuss technicalities associated with its optimisation. At its essence, our focus is on obtaining an alternative factorisation of the class conditional densities to that used in \nb. We address this problem by formulating it as a projection pursuit, in which we seek a {\em projection matrix}, $\V \in \R^{p\times \pp}$, for which the factorisation of the estimated densities of the random variables $\V^\top X | Y = k$; $k \in [K]$, leads to improved classification accuracy over that arising from the factorisation of the estimated densities of $X|Y=k$; $k \in [K]$, as in the standard \nb~approach. An added benefit of treating this as a projection pursuit is the possibility for using it to perform dimension reduction, i.e., when $\pp < p$. The objective we use to optimise the projection matrix is the multinomial likelihood, given by
\begin{align}\label{eq:classlik}
\mathcal{L}(\V) \propto \prod_{i=1}^n  \frac{\hat \pi_{y_i}\hat f_{\V^\top X | Y=y_i}(\V^\top \x_i)}{\sum_{k=1}^K\hat \pi_{k}\hat f_{\V^\top X | Y=k}(\V^\top \x_i)}.
\end{align}
Existing projection pursuit methods in the literature which utilise this multinomial likelihood include those based using Gaussian densities~\citep{hofmeyr2020optimal} and  Gaussian mixtures~\citep{peltonen2005discriminative} for the class densities.

\subsection{Optimising $\mathcal{L}(\V)$}

As is common, we optimise the logarithm of the likelihood, which is thus given by (up to an additive constant)
\begin{align}\label{eq:loglik}
    \ell(\V) =& \sum_{i=1}^n\Bigg( \log\left(\hat \pi_{y_i} \hat f_{\V^\top X|Y=y_i}(\V^\top \x_i)\right) - \log\left(\sum_{k=1}^K \hat \pi_k \hat f_{\V^\top X|Y=k}(\V^\top \x_i)\right)\Bigg),
\end{align}
where with the \nb~factorisation each estimated class density is given by
\begin{align}
\hat f_{\V^\top X|Y=y_i}(\V^\top \x) = \prod_{d=1}^\pp \hat f_{\V_d^\top X|Y=y_i}(\V_d^\top \x),     
\end{align}
with $\V_d$ the $d$-th column of $\V$. The factorisation allows us to use efficient methods for univariate kernel density estimation to evaluate the objective (and its gradient)~\citep{hofmeyr2019fast, hofmeyrFKSUM}. In particular, we set
\begin{align}\label{eq:kde}
    \hat f_{\V^\top_d X|Y = k}(x) = \frac{1}{n_k h} \sum_{i: y_i = k}K\left(\frac{x - \V^\top_d \x_i}{h}\right),
\end{align}
where $K$ is a kernel function and $h > 0$ is the bandwidth. 
Now, it is worth pointing out that if the projections of the observations along each of the columns of $\V$ are all distinct, then the objective in Eq.~(\ref{eq:loglik}) approaches its maximum value of 0 simply by multiplying $\V$ with an arbitrarily large constant. This is because, if the bandwidth is fixed, the kernel evaluations in Eq.~(\ref{eq:kde}) all approach zero except when $x$ is equal to one of the projections, and hence would have, for each $d \in [p]$,
\begin{align*}
    \lim_{\alpha \to \infty} \hat f_{\alpha\V^\top_d X|Y = k}(\alpha \V^\top_d \x_i) &= \left\{\begin{array}{ll}
        \frac{1}{n_k h}K(0), & k = y_i \\
        0, & \mbox{otherwise}.
    \end{array}\right.
\end{align*}

Although it is possible to directly constrain the magnitude of $\V$ in order to prevent this, we instead introduce a penalty to the objective in order to resolve the ``collapse'' described above. Specifically, we focus on the optimisation problem given by
\begin{align}\label{eq:penalised_objective}
    \max_{\V \in \R^{p\times p'}}\frac{1}{n}\ell(\V) - \lambda ||\V||_F^2,
\end{align}
where $\lambda > 0$ is the strength of the penalty and $||\cdot||_F$ is the Frobenius norm, and simply set the bandwidth equal to 1.

We prefer this approach to the constrained formulation, as optimising the projection matrix without fixing its magnitude, while using a fixed bandwidth, has a similar effect to fixing the scale of the matrix but optimising the bandwidth. Computationally, however, the former is far preferable since there are fewer parameters to be optimised, and no constraints on the objective. Penalising the magnitude of $\V$ then simply mitigates the overfitting effect which would result from the ``perfect fit'' as the bandwidth tends to zero relative to the scale of the projected data. 
%

%
%
%

Now, it will be convenient going forward, for the sake of brevity, to introduce the following notation. We will use $Z$ to represent the projection of the random variable, $\V^\top X$, and $\z_i$ or $\z$ to represent the projected data point $\V^\top \x_i$, or an arbitrary realisation, $\V^\top \x$, respectively. To shorten class conditioning notation $\cdot | Y = j$, we will henceforth simply write $\cdot |j$, so that, in total, when there is no ambiguity about the value of $\V$, we can write $\hat f_{Z|j}(\z)$ for $\hat f_{\V^\top X | Y = j}(\V^\top \x)$. In addition, we will use the subscript $\bar t$ to mean ``all-but-the-$t$-th'', so that $Z_{\bar t} = (Z_1, ..., Z_{t-1}, Z_{t+1}, ..., Z_\pp)$, and similarly $\z_{i\bar t} =$\\ $(z_{i1}, ..., z_{i(t-1)}, z_{i(t+1)}, ..., z_{i\pp})$. This is particularly useful in the context of factorising densities, e.g., $\hat f_{Z|j}(\z) = \hat f_{Z_{\bar t}|j}(\z_{\bar t})\hat f_{Z_t|j}(z_t)$. Finally, we set $\hat f_Z = \sum_{j=1}^K \hat \pi_j \hat f_{Z|j}$.

In order to derive an expression for the gradient of $\ell(\V)$, we will write
\begin{align}
    \ell(\V) &= \tilde \ell(\Z),\\
    \tilde \ell(\Z) &:= \sum_{i=1}^n\left( \log\left(\hat \pi_{y_i} \hat f_{Z|y_i}(\z_i)\right) - \log\left(\sum_{k=1}^K \hat \pi_k \hat f_{Z|k}(\z_i)\right)\right),\\
    \Z &:= \X\V.
\end{align}
%
%
Now, since the effect of varying the $t$-th column of $\V$ only affects the $t$-th column of $\Z$,
we can then evaluate the vector of partial derivatives of $\ell(\V)$, with respect to the elements in the $t$-th column of $\V$, using the chain rule decomposition
\begin{align}
    \nabla_{\V_t}\ell(\V) = \nabla_{\Z_t}\tilde\ell(\Z) D_{\V_t} \Z_t,
\end{align}
where $D_{\V_t} \Z_t$ is the matrix with $i,j$-th entry $\frac{\partial z_{it}}{\partial \V_{jt}}$, and is simply equal to $\X$. 
%
%
In order to evaluate the elements of $\nabla_{\Z_t} \tilde \ell(\Z)$, consider that
\begin{align*}
    \frac{\partial \tilde \ell(\Z)}{\partial z_{st}} =& \sum_{i=1}^n \bigg(\frac{\frac{\partial}{\partial z_{st}}\hat f_{Z|y_i}(\z_i)}{\hat f_{Z|y_i}(\z_i)} - \frac{\frac{\partial}{\partial z_{st}}\hat f_{Z}(\z_i)}{\hat f_{Z}(\z_i)}\bigg)\\
    =& \sum_{i=1}^n \bigg(\frac{\frac{\partial}{\partial z_{st}} \prod_{d = 1}^{\pp}\hat f_{Z_d|y_i}(z_{id})}{\prod_{d = 1}^{\pp}\hat f_{Z_d|y_i}(z_{id})}- \frac{\sum_{j=1}^K \hat \pi_j \frac{\partial}{\partial z_{st}}\prod_{d = 1}^{\pp}\hat f_{Z_d|j}(z_{id})}{\hat f_{Z}(\z_i)}\bigg)\\
    =& \sum_{i:y_i=y_s} \frac{\frac{\partial}{\partial z_{st}} \hat f_{Z_t|y_s}(z_{it})}{\hat f_{Z_t|y_s}(z_{it})} - \sum_{i=1}^n\frac{\sum_{j=1}^K \hat \pi_j \hat f_{Z_{\bar t}|j}(\z_{i\bar t})\frac{\partial}{\partial z_{st}}\hat f_{Z_t|j}(z_{it})}{\hat f_{Z}(\z_i)},
\end{align*}
where the first sum in the final expression only includes those observations from the same class as $\x_s$ since changes to the value of $\z_s$ will not affect the estimated densities of other projected points evaluated in {\em their} class densities. To obtain an explicit expression for these partial derivatives, we proceed with the simpler first term to begin;
\begin{align*}
    T_1 :=& \sum_{i:y_i=y_s} \frac{\frac{\partial}{\partial z_{st}} \hat f_{Z_t|y_s}(z_{it})}{\hat f_{Z_t|y_s}(z_{it})}\\
    =& \sum_{\substack{i:y_i=y_s\\i\not = s}} \frac{\frac{\partial}{\partial z_{st}} \hat f_{Z_t|y_s}(z_{it})}{\hat f_{Z_t|y_s}(z_{it})} + \frac{\frac{\partial}{\partial z_{st}} \hat f_{Z_t|y_s}(z_{st})}{\hat f_{Z_t|y_s}(z_{st})}\\
    =& \sum_{\substack{i:y_i=y_s\\i\not = s}} \frac{1}{\hat f_{Z_t|y_s}(z_{it})} \frac{1}{n_{y_s}}\sum_{j:y_j=y_s}\frac{\partial}{\partial z_{st}}K\left({z_{jt}-z_{it}}\right) + \frac{1}{\hat f_{Z_t|y_s}(z_{st})} \frac{1}{n_{y_s}}\sum_{j:y_j=y_s}\frac{\partial}{\partial z_{st}}K\left({z_{jt}-z_{st}}\right) \\
    &= \frac{1}{n_{y_s}}\sum_{i:y_i=y_s} K'\left({z_{st}-z_{it}}\right)\left(\frac{1}{\hat f_{Z_t|y_s}(z_{it})} + \frac{1}{\hat f_{Z_t|y_s}(z_{st})}\right).
\end{align*}
Note that $K'\left({z_{st}-z_{st}}\right) = 0$, and hence both sums can be taken over all $i: y_i = y_s$. Now, with the explicit form of the terms $\hat f_{Z_t|j}(z_{it})$, the negative of the second term in the partial derivative is equal to
\begin{align*}
-T_2 :=& \sum_{i=1}^n \frac{1}{\hat f_Z(\z_i)}\sum_{j=1}^K\hat \pi_j \hat f_{Z_{\bar t}|j}(\z_{i\bar t})\frac{1}{n_j} \sum_{l:y_l=j}\frac{\partial}{\partial z_{st}}K\left({z_{lt}-z_{it}}\right),
\end{align*}
where the non-zero partial derivatives occur only if either index $i$ or $l$ is equal to $s$. We thus have,
\begin{align*}
    -T_2 =& \frac{1}{\hat f_Z(\z_s)}\sum_{j=1}^K\frac{\hat f_{Z_{\bar t}|j}(\z_{s\bar t})}{n}\sum_{l:y_l=j} K'\left({z_{st}-z_{lt}}\right) + \sum_{i=1}^n \frac{1}{\hat f_{Z}(\z_i)} \frac{\hat f_{Z_{\bar t}|y_s}(\z_{i\bar t})}{n}  K'\left({z_{st}-z_{it}}\right)\\
    =& \frac{1}{n}\sum_{i=1}^n\Bigg(\frac{\hat f_{Z_{\bar t}|y_i}(\z_{s\bar t})}{\hat f_Z(\z_s)} K'\left({z_{st}-z_{it}}\right) + \frac{\hat f_{Z_{\bar t}|y_s}(\z_{i\bar t})}{ \hat f_Z(\z_i)} K'\left({z_{st}-z_{it}}\right)\Bigg).
\end{align*}
  
These expressions for the terms $T_1$ and $T_2$ are now in the form described by~\cite{hofmeyr2019fast}, for which computation for all $s$ is possible in $\mathcal{O}(n \log n)$ time.

Finally, to optimise the objective $\frac{1}{n}\ell(\V) - \lambda ||\V||^2_F$ we then use the quasi-Newton Limited memory BFGS algorithm~\citep{LBFGS}.

\subsection{Other Practicalities}

\subsubsection{Discrete covariates:} \label{sec:OHE}

One of the appealing properties of the standard \nb~model is its ability to easily incorporate both continuous and discrete covariates, whereas in general modelling multivariate discrete covariates or combinations of discrete and continuous covariates in a probabilistic classification model, i.e., one which explicitly models the class conditional distributions, might be extremely challenging. This is especially important when the discrete covariates are categorical and the ordering of and relative differences between the values with which they are encoded may be arbitrary.

One of the most common approaches for incorporating categorical covariates is one-hot-encoding (OHE), in which a covariate taking $G$ distinct values is replaced with $G-1$ binary (0/1) variables, and is then treated as a numerical variable. Although this alleviates the problem of arbitrary differences and ordering; problems of discreteness remain, especially if some of the resulting binary variables are constant within some classes.

In the projection pursuit framework where the projection matrix, $\V$, is dense (contains no zeroes), all of the random variables $\V_d^\top X; d = 1, ..., p'$ are continuous as long as there is even a single covariate whose support is the entire real line. Furthermore, even if there are no such continuous covariates, dense projections of multiple binary variables arising from OHE will often assume a large number of distinct values and so modelling them as continuous often leads to reliable estimation of classification models.

It is worth noting that this benefit does not apply to models which are invariant to rotation of covariates (as are many classification models) since the joint distribution of $\V^\top X$ is not continuous even if each of its marginal distributions is. Within the OPNB formulation the random variables $\V_d^\top X|Y = k; d \in [p'], k \in [K]$ are modelled separately before being combined, and hence our model is comparatively robust to the presence of many categorical variables. We briefly discuss this in relation to our experimental results in Section~\ref{sec:experiments}.


\paragraph{Initialisation:} The objective function given in Eq.~(\ref{eq:penalised_objective}) is non-concave and so initialisation of $\V$ can have a significant impact on the quality of the resulting model. We have experimented with a number of heuristics for initialisation, with none obviously standing out as consistently superior to others over many contexts. For simplicity and for the sake of reproducibility, in our experiments we only use a single initialisation in each experiment, given by the leading principal component directions of the data. However, in any single application it is worthwhile exploring the models arising from multiple initialisations. As mentioned previously, one of the benefits of the projection pursuit framework is that it leads naturally to visualisations of what the model has captured, via scatter-plots of the projected data. These plots can also sometimes be used as diagnostics in order to ascertain whether any transformations of the data may be beneficial. We discuss this in greater depth in Section~\ref{sec:experiments}.

\subsection{Connection to Class Conditional Independent Component Analysis}

Independent Component Analysis~\citep[ICA]{comon_ica}
is a projection pursuit problem that seeks the projection matrix, $\V$, which minimises a measure of statistical dependence in the elements of $\V^\top X$, where $X$ is the random variable assumed to underlie ones observations. A popular measure of dependence in ICA is the mutual information. It can be shown that an equivalent problem is to minimise the sum of the differential entropies of the random variables $\V^\top_1 X, ... , \V^\top_{\pp} X$, i.e., to minimise
\begin{align}\label{eq:ica}
\sum_{d=1}^\pp E\left[-\log\left(f_{\V_d^\top X}(\V_d^\top X)\right)\right].    
\end{align}
There is an inherent and intuitive connection between ICA and \nb, since the factorisation used in \nb~treats the random variables $X|Y=k; k\in [K]$, as having independent entries. Indeed, as we outlined breifly in the introduction, a variation of non-parametric discriminant analysis applies ICA to the subsets of the data arising in each of the classes, before using a \nb~factorisation in order to estimate the class densities~\citep{CCICA}. This is an appealing idea for its simplicity, and the method has shown some success. However, as the projection matrices are obtained separately for each class, there is no reason to expect any of these will be useful for the discrimination of classes. In other words, this method is motivated by obtaining more accurate estimates of the class densities by combining ICA with \nb~factorisation, but the estimation procedure is fully agnostic of the actual classification objective.

Let us now turn our attention to the log-likelihood objective we consider, given in Eq.~(\ref{eq:loglik}). The first term is given by
\begin{align*}
    \sum_{i=1}^n &\log\left(\hat \pi_{y_i} \hat f_{\V^\top X|y_i}(\V^\top \x_i)\right) = \sum_{k=1}^K\left(n_k \log(\hat \pi_k) + \sum_{i:y_i = k} \sum_{d=1}^\pp \log\left(\hat f_{\V^\top_d X|k}(\V^\top_d \x_i)\right)\right)\\
    =& C + \sum_{k=1}^K n_k \sum_{d=1}^\pp \frac{1}{n_k}\sum_{i:y_i = k} \log\left(\hat f_{\V^\top_d X|k}(\V^\top_d \x_i)\right)\\
    =& C + \sum_{k=1}^K n_k  \sum_{d=1}^\pp \overline{\log\left(\hat f_{\V^\top_dX|k}(\V^\top_d \x)\right)},
\end{align*}
where $\overline{\log\left(\hat f_{\V^\top_dX|k}(\V^\top_d \x)\right)}$ is supposed to represent the empirical average of the estimated log density for class $k$, and $C$ is a constant independent of $\V$. By replacing the expectations in the ICA objective, Eq.~(\ref{eq:ica}), with their empirical estimates, we can see that the first term in our objective is simply the weighted sum of the ICA objectives for the subsets of the data belonging to each class (recall that our objective is to maximise the log-likelihood, whereas ICA minimises the differential entropy, or equivalently maximises the negative entropy). This is appealing since this part of the objective encourages our solution to have low dependence in the elements of the random variables $\V^\top X|Y=k; k\in [K]$, and hence the \nb~factorisation is likely to lead to relatively accurate estimation of the class densities. While this is a desirable property, as previously discussed, the discrimination of classes does not yet factor in. The negative of the second term in our objective is given by
\begin{align*}
    \sum_{i=1}^n \log\left(\sum_{k=1}^K \hat \pi_k \hat f_{\V^\top X|k} (\V^\top \x_i)\right).
\end{align*}
This term measures the estimated total (mixture) log-likelihood of the projected points. This can be thought of as a penalty which discourages projection matrices, $\V$, upon which the projected points are fit well by the density which incorporates all of the classes. In other words the first term encourages a good fit of the points in their own classes, while the second term ensures points aren't simply well explained by all of the classes as this would not allow strong discrimination.

\section{Experiments}\label{sec:experiments}

In this section we explore the classification accuracy of the proposed approach in comparison with relevant alternatives, on a large collection of publicly available benchmark data sets. We also perform an investigation into the relationships between the classification performance of the proposed approach and the characteristics of the data to which it is applied. 

\subsection{Data Sets and Preprocessing}

For our experiments, we considered all 162 classification data sets in the Penn Machine Learning Benchmarks repository~\citet{citePMLB}. We conducted the following preprocessing policy, which was executed in order:
\begin{enumerate}
\item 
For data sets containing more than 3000 samples we performed stratified sampling to obtain a sample of size 3000 which (approximately) respects the class proportions in the complete data set.
\item 
We removed classes with fewer than 10 observations from the data.
\item 
Covariates with zero sample standard deviation were excluded from the experiments.
\item 
Covariates with at most 5 unique values were treated as categorical and one-hot encoded.
\item 
Small Gaussian perturbations were added to the data to avoid numerical issues, arising when variables have zero standard deviation within one of the classes. The standard deviation of the perturbations added to a covariate was equal to one percent of the standard deviation of the covariate itself.
\item 
If a data set contains more than 300 covariates then we replace them with their first 300 principal components.
\end{enumerate}

\subsection{Classification models, Tuning, and Evaluation}

Our focus is mainly on alternative probabilistic classification models, i.e., those using applications of Bayes' rule applied to posterior probabilities of class membership arising from estimated class conditional distributions. But for context we also included the popular Support Vector Machine~\citep[SVM]{SVM}.
Below is a complete list of the models we considered, along with their respective tuning parameters.
\begin{enumerate}
    \item \nb~(NB): The \nb~model with class conditional marginal densities estimated using KDE. Note that using standard bandwidth selection rules for KDE is not appropriate for discrete covariates. When tuning bandwidths we therefore separate the OHE encoded categorical covariates from the continuous 
    ones and tune a single bandwidth $\alpha \in \{0.1, 0.2, ..., 0.5\}$ to be used for all classes and all binary variables, and a single multiplication factor, $\gamma \in \{1/3, 1/2, 1, 2, 3\}$, and set the bandwidth for a pair of class and continuous covariate equal to $\gamma$ multiplied by Silverman's rule of thumb value~\citep{silverman2018density} for the corresponding subset of points. Note that using an over-smoothing bandwidth on OHE encoded categorical variables is very similar in the resulting probabilities to applying a Laplace adjustment to the empirical proportions in the different values of the categorical variable(s). 
    %
    
    Note that we also considered the Gaussian \nb~model, in which the class conditional marginals are treated as Gaussian, but the performance of this approach was overall worse than all other methods to the extent that its inclusion made the comparison between other methods more challenging, and so we omit these results.
    \item Class Conditional ICA + \nb~(CCICA): The method of~\cite{CCICA} in which class conditional densities are estimated from the product of the marginal density estimates of the ICA transformed classes. We tuned both the (shared) dimension of the ICA transformations for the classes, from the set $[\min\{p, 20\}]$; and also a bandwidth multiplier for the KDE estimates of the marginal densities, using the same approach as for NB. Note that the issues of handling one-hot-encoded variables using non-parametric density estimates is mitigated in the same way by CCICA as in the proposed approach and so it is not necessary to modify these density estimates for discrete covariates (nor is it clear how this could be appropriately done).
    \item Kernel Density Discriminant Analysis (KDDA): The probabilistic classification model in which class densities are estimated with a multivariate kernel estimator. To accommodate the continuous as well as one-hot-encoded variables we used diagonal bandwidth matrices for each class, and tuned hyperparameters $\alpha$ and $\gamma$ from the same sets as for \nb.
    The bandwidth matrix for class $k$ was then, assuming w.o.l.o.g. that the continuous covariates lie in the leading columns of the data matrix, set to
    \begin{align*}
        \left[\begin{array}{cc}
            \gamma\left(\frac{4}{n_k(p_{c}+2)}\right)^{\frac{1}{p_{c}+4}} \Delta(\hat \Sigma_k)^{1/2} & \mathbf{0}\\
            \mathbf{0} & \alpha \mathbf{I}
        \end{array}\right],
    \end{align*}
    where $\Delta(\hat\Sigma_k)$ is $\hat \Sigma_k$, the sample covariance matrix from class $k$, but with all off diagonal elements set to zero. Note that the factor $\left(\frac{4}{n_k(p_{c}+2)}\right)^{\frac{1}{p_{c}+4}}$, where $p_c$ is the number of continuous covariates, arises from Silverman's rule of thumb when using a Gaussian kernel as we do for KDDA.
    \item Linear Discriminant Analysis (LDA): The probabilistic classification model in which class conditional densities are treated as Gaussian, and the maximum likelihood estimate for a shared covariance matrix is used. We tuned the number of discriminant dimensions to use for classification.
    \item Regularised Discriminant Analysis (RDA): The probabilistic classification model with Gaussian class densities in which class $k$ is given covariance matrix $\lambda \hat\Sigma_k + (1-\lambda)\hat\Sigma_W$, where $\hat \Sigma_k$ is the maximum likelihood estimate of the covariance of class $k$ and $\hat\Sigma_W$ is the shared covariance matrix used in LDA. Tuning over $\lambda \in [0, 1]$ traverses the spectrum of complexity joining LDA for $\lambda = 0$ to Quadratic Discriminant Analysis (QDA) for $\lambda = 1$. We considered values for $\lambda$ in $\{0.1,0.2,\hdots,1 \}$.
    %
    \item Optimal Projections for Gaussian Discriminants (OPGD): A recent method based on classification under a Gaussian discriminant model fit to an optimal projection of the data, as determined by the same multinomial likelihood objective we employ. We tuned only the dimension of the projection, from the set $[\min\{p,20\}]$.
    \item Support Vector Machine (SVM): The kernelised linear classifier based on the penalised hinge-loss objective.     
    We used a Gaussian kernel parameterised as $K(\x_1, \x_2) = \exp(-\alpha ||\x_1-\x_2||^2)$, and tuned $\alpha$ over $\{2^{2i+1}| i = -8, -7, ..., 1\}$ and the ``cost'' parameter, which balances the trade-off between the ridge-like penalty term and the hinge-loss objective, from the set $\{2^{2i+1} | i = -3, -2, ..., 7\}$. This collection of values is recommended internally by libsvm \citep{citelibsvm} and we used this implementation in our experiments. In libsvm, multi-class classifiers are constructed from binary SVM classifiers in a ``one-versus-one'' manner.
    \item Optimal Projections for \nb~(OPNB): The proposed method. We tuned $\lambda$ over $\{10^{-4}2^i|i = 0, 1, ..., 9\}$, and for simplicity only used a single setting of $p' = \min\{p, 20\}$. Of course it is possible to also tune over $p'$, however we have found the penalisation approach effective in regularising over-parameterisation inherent in setting $p'$ very large. We have also found no examples where using a greater setting of $p'$ leads to appreciably better performance. In the interest of computational speed we also used binning with 1000 bins to further speed up the kernel computations needed to fit OPNB models.

\end{enumerate}

\noindent
In order to select appropriate tuning parameters we estimated the misclassification rate for each combination using 5-fold cross-validation applied on a training set comprising 75\% of a given data set. We then recorded the misclassification rate on the remaining 25\% test data, from the model(s) trained on the complete training set with their selected hyperparameters. To mitigate the effects of randomness, we repeated each split into training/test sets 10 times and used the same cross-validation folds for all methods. We also performed stratified sampling at every stage to approximately preserve the class proportions in every test set and cross-validation fold. Because some of the methods applied (including our own) are not scale invariant, for every training instance we scaled the data, dividing each variable by the standard deviation of the corresponding observations within the training set.

\subsection{Classification Performance}

In this subsection we present summaries of the overall performance of all 8 methods across the 162 data sets. Note that when combining results across data sets of varying characteristics it is important to standardise the performance metric(s), to make them comparable across different data sets. This is because the accuracy achievable across different data sets may differ to the extent that it overshadows the differences in performance between different methods compared on the same data set.

We apply two standardisations to the average test errors from all eight methods on each data set. Specifically, if $Err(M_{i,d})$ is the average test misclassification rate for method $i$ from data set $d$, then we consider
\begin{align*}
    Err^*(M_{i,d}) :=& \ \frac{Err(M_{i,d})-\min_j Err(M_{j,d})}{1-\min_j Err(M_{j,d})} & \mbox{ min-normalisation}\\
    Err^{**}(M_{i,d}) :=& \ \frac{Err(M_{i,d}) - \overline{Err(M_{d})}}{S(Err(M_{d}))} & \mbox{ studentisation},
\end{align*}
where $\overline{Err(M_{d})}$ and $S(Err(M_{d}))$ are the mean and standard deviation of the average misclassification rates of all methods applied to data set $d$.


\begin{figure*}
    \subfigure[Min-normalised Errors]{\includegraphics[width = .49\textwidth]{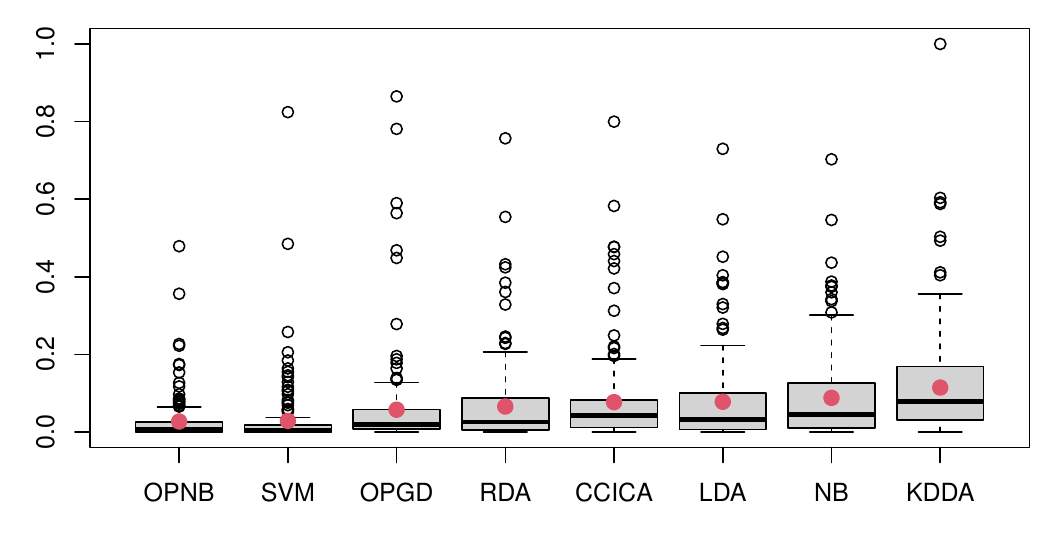}\label{fig:boxplot_sderr1}}
    \subfigure[Studentised Errors]{\includegraphics[width = .49\textwidth]{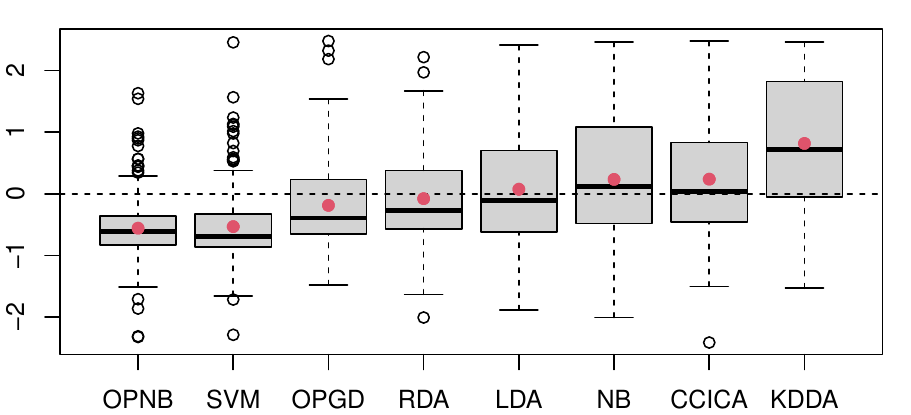}\label{fig:boxplot_sderr2}}
    \caption{Boxplots of standardised error distributions for all methods across the 162 data sets, using min-normalisation and studentisation. The red dots correspond with the averages in each case, and the box-plots have been ordered based on increasing average values. 
    \label{fig:boxplot_sderr}}
\end{figure*}



\subsubsection{Error Distributions}

Boxplots of the distributions of standardised errors achieved by the different methods across all 162 data sets are shown in Figure~\ref{fig:boxplot_sderr}. In addition the averages are shown with red dots. The methods have been ordered based on average standardised error. OPNB and SVM have extremely similar average performance based on both standardisations, with OPNB having slightly lower average. Although the median performance of SVM is slightly lower than that of OPNB, SVM has a longer upper tail leading to slightly higher average. 
%
%
Among the other methods OPGD and RDA have similar overall performance, but with OPGD slightly outperforming RDA on average across all the data sets considered. LDA, NB and CCICA have quite similar performance to one another on average, while KDDA is considerably worse than any of the other methods we considered. It is noteworthy that the performance of OPNB and SVM is, overall, substantially better than the other methods under comparison. This is interesting since, as mentioned previously, SVM is the only non-probabilistic classifier we have used, and yet the performance of OPNB much more closely aligns with SVM than with other probabilistic models.

It is also noteworthy that, across many contexts, whether the modification to the standard \nb~model used in CCICA is actually beneficial is questionable. Indeed the average performance is similar and, if anything, the CCICA model lends itself less to interpretation than does the standard \nb~model.\\
%
%
\\
Now, it is of course possible to use other standardisations of the raw misclassification rates, such as mapping the error rates for all methods to span the interval $[0, 1]$ exactly as described by~\cite{ESL}, however we prefer the similar min-normalisation described above as it is not sensitive to the performance of the worse performing methods. For the readers' interest the ordering of the average error based on this standardisation is the same as for min-normalisation except the positions of LDA and CCICA are swapped. The ordering based on the raw misclassification rates is the same as that arising from the $[0, 1]$ mapping.


\begin{table*}[h]
    \centering
    \begin{tabular}{r|p{1.5cm}p{1.5cm}p{1.5cm}p{1.5cm}p{1.5cm}p{1.5cm}p{1.5cm}p{1.5cm}}
   &OPNB&SVM&OPGD&RDA&CCICA&LDA&NB&KDDA\\
   \hline
OPNB& &75&109&111&130&114&120&140\\
SVM&79& &112&109&126&107&119&145\\
OPGD&44&45& &85&105&87&95&127\\
RDA&49&48&73& &87&82&103&115\\
CCICA&29&32&53&72& &78&80&105\\
LDA&44&52&69&72&82& &94&110\\
NB&41&43&67&58&81&67& &102\\
KDDA&21&17&35&46&57&50&58&

    \end{tabular}
    \caption{Comparative performance of pairs of methods. Each cell in the table contains the number times the method listed row-wise outperformed the method listed column-wise. For example OPNB achieved lower average misclassification rate than SVM on 75 data sets, and SVM outperformed OPNB on 79. On the remaining $8 = 162-(75+79)$ data sets they achieved the same average misclassification rate.}
    \label{tab:winsnew}
\end{table*}

\subsubsection{Pair-wise Performance}

Although the boxplots allow us to determine relative performances of different methods in general, they do not directly imply any relative comparisons on any individual data sets. Table~\ref{tab:winsnew} summarises the pairwise comparative performances for all methods. Each cell in the table shows the number of times the method listed row-wise outperforms the method listed column-wise. Although OPNB has the lowest average misclassification rate across all data sets, SVM has lower error than OPNB more often that the reverse.

Ignoring instances of equal error (ties), the proposed approach outperforms SVM approximately 49\% of the time, and outperforms each other method in at least 60\% of cases.


\subsection{Dissecting Classification Performance of OPNB}

Here we perform a brief investigation into the relationships between the characteristics of the different data sets and the performance of OPNB. Specifically, we characterised each data set by six statistics, which we describe below, and then regressed the standardised misclassification rate\footnote{We used the studentised performance as its distribution is closer to Gaussian than that of the min-normalised performance and, all other things being equal, more likely to suitably used in an OLS model.} of OPNB on these statistics, using a simple ordinary linear model. The reason for using a standardised performance metric is it allows us to investigate the extent of the relationships between data set characteristics and performance of OPNB \textit{relative to the performance of other relevant models}. From a model selection perspective, this is arguably more instructive than the relationships with the raw performance values.

The statistics used to characterise each data set are:
\begin{enumerate}
    \item Number of dimensions ($p$): The total number of dimensions after one-hot-encoding.
    \item Number of observations ($n$): Recall that we subsampled large data sets down to a maximum of 3000 due to the large computation time required to tune and train all methods on such a large number of data sets.
    \item Proportion of categorical variables: The proportion of binary variables in a data set after one-hot-encoding. The reason for taking the number of binary variables rather than the number categorical variables in the original data set is that this accounts for the granularity (number of distinct values) of the categorical variables as well as their number.
    \item Class imbalance: The variance of the class proportions. 
    \item Number of classes ($K$): The total number of classes.
    \item Complexity of decision boundaries: This is an artificial score designed to capture the degree of complexity of the class decision boundaries, by comparing the performance of the nearest centroid (NC) classifier and the one-nearest-neighbour (1-NN) classifier. While the 1-NN classifier can accommodate highly non-linear decision boundaries (at the cost of high variance), the NC classifier is very stable (has low variance) but has rigid linear decision boundaries. Although neither is likely to perform well in general, they arguably occupy two ends of the spectrum of complexity and their relative performance can be used to capture to some extent the complexity of the class decision boundaries. Importantly they are also separate from the set of classifiers used for comparison.
    To determine the actual complexity score we compute the min-normalised misclassification rates using only these two models, and take as the complexity score the standardised error of NC. When the performance of NC is substantially worse than that of 1-NN this suggests a potentially complex decision boundary, and this is precisely when the standardised error of NC will be large.
\end{enumerate}

To make the regression coefficients more comparable with one another directly, we standardised the values of these statistics across data sets to each have unit variance. Note that we prefer using the regression coefficients over the marginal correlations as they take account of the other characteristics of the data sets. For example, we would expect that, all other things being equal, increasing the number of observations will improve the performance of all methods, however within the data sets considered it is possible that all of the large data sets had other characterisics which led to poorer performance by OPNB and the regression coefficients are able to partly account for such a phenomenon. 

The only positive regression coefficient, i.e. one whose increase is associated with an increase in the error of OPNB, is that associated with the number of classes. However, the magnitude of the coefficient is very small, as is that for the total number of dimensions. The strongest relationships, as captured by this simple model, are those associated with the proportion of categorical variables (the strongest by far); the complexity of the decision boundaries; and the sample size.

We applied the same simple analysis to the other classification models, and found that the performance of OPNB has an almost three-fold stronger association with the proportion of categorical variables than any other model (after accounting for the scale of the standardised performances across different models), and that only SVM had a stronger association with sample size than OPNB. Recall that because the performances have been standardised, a stronger association with sample size indicates that the associated model is, in a sense, better able to leverage larger samples than the other methods considered.\\
\\
Although simplistic, this investigation provides some evidence that OPNB is a flexible classifier which utilises larger samples effectively, but is robust to the presence of large numbers of categorical covariates.



%


\subsection{Projection Plots and Diagnostics}

As mentioned previously one major benefit of the projection pursuit framework is the fact it naturally leads to instructive and informative visualisations, by means of scatter plots of the projected data. This can help identify whether simple modifications; to the data, the implementation of the method, or its hyperparameters, may be useful in obtaining an improved model.
In this subsection we briefly explore a few cases, and focus on instances where the performance of OPNB was comparatively poor in order to illustrate some simple modifications and/or changes which can lead to improved performance.

\subsubsection{Highly Elongated Covariance Matrix}

\begin{figure}
    \centering
    \subfigure[Hill-Valley]{\includegraphics[width=0.6\linewidth]{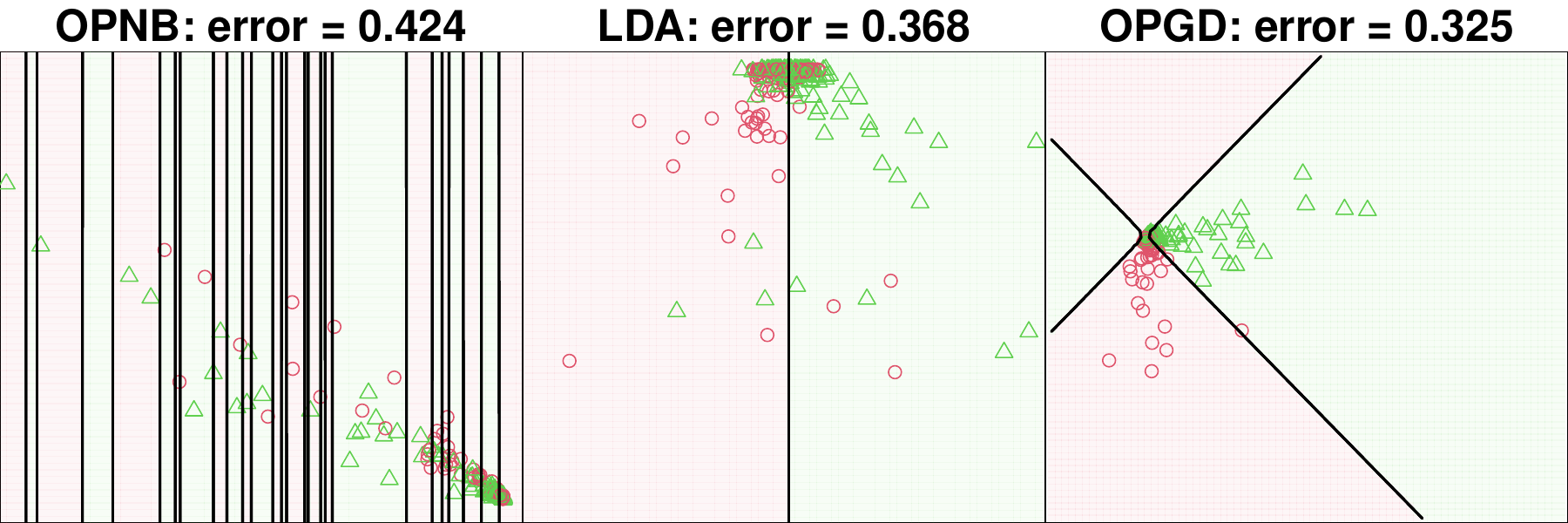}}
    \subfigure[Hill-Valley with Noise]{\includegraphics[width=0.6\linewidth]{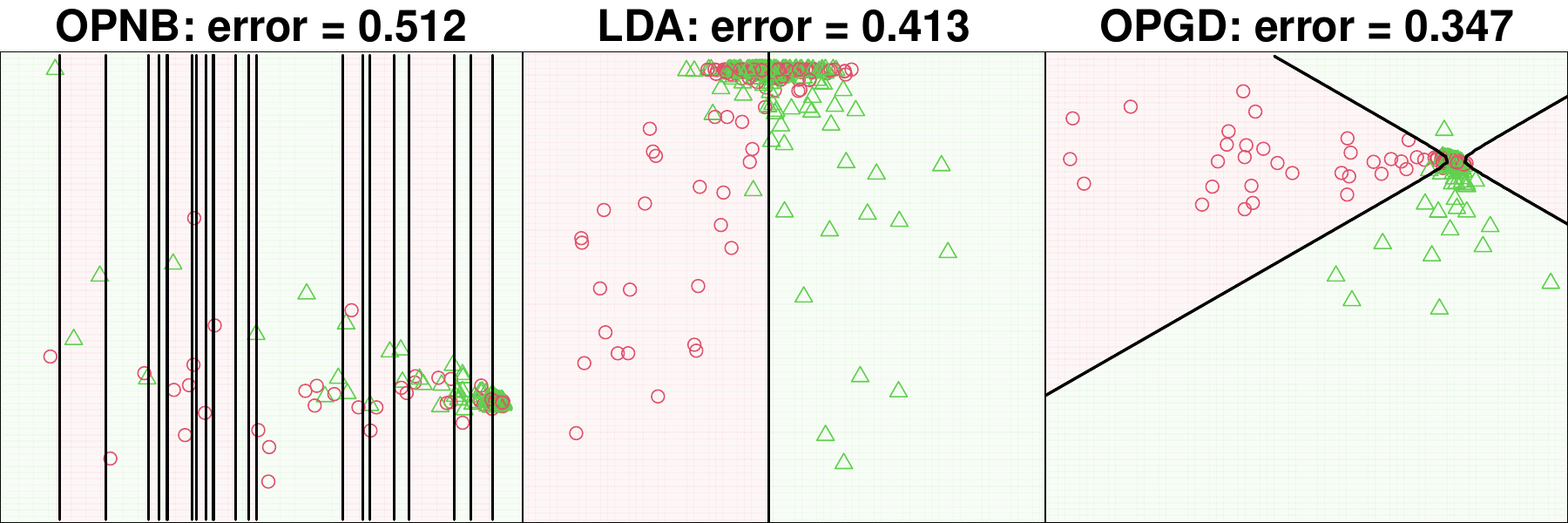}}
    \caption{Projection plots of the Hill-Valley data sets from OPNB, LDA and OPGD. The OPNB plots show overfitting caused by the existence of directions in which the scale of the data is extremely high, due to the strong correlation between covariates.}
    \label{fig:hillvalley}
\end{figure}

The ``Hill-Valley'' data sets~\citep{hillValley} are synthetic data sets in which the 100 covariates are sequential, and each instance either shows a peak/hill or a trough/valley, depending on its class. The high dependence between the covariates leads to poorly conditioned covariance matrices in which the first principal component accounts for more than 99\% of the total variance. The best two-dimensional projections based on OPNB, LDA\footnote{since there are only two classes LDA only has one non-arbitrary discriminant direction, and so in order to produce a two-dimensional visualisation we used as the second the first principal component of the data projected into the null-space of the LDA discriminant subspace.} and OPGD are shown in Figure~\ref{fig:hillvalley}. In each plot the lines show the decision boundaries from each of the models, while the points are projections of an independent test set.

Although none in the suite of classification models we have considered for our experiments is designed to handle sequential data, we nonetheless find this example instructive as it highlights a potential limitation with our basic implementation of OPNB. The penalisation of the magnitude of the projection matrix is designed to mitigate the overfitting effect of the non-parametric density estimates when the scale of the projected data is very large relative to the fixed bandwidth of 1. However, when there is a single (or few) directions in which the scale of the data is extremely large due to high correlation between the covariates, this simple penalisation is ineffective. This is clearly visible in Figure~\ref{fig:hillvalley}, where the decision boundaries induced by the OPNB model are caused by density estimates along the horizontal direction which have a very large number of modes, even picking out the majority of singletons in the tails.

\begin{figure}
    \centering
    \subfigure[Hill-Valley]{\includegraphics[width=0.48\linewidth]{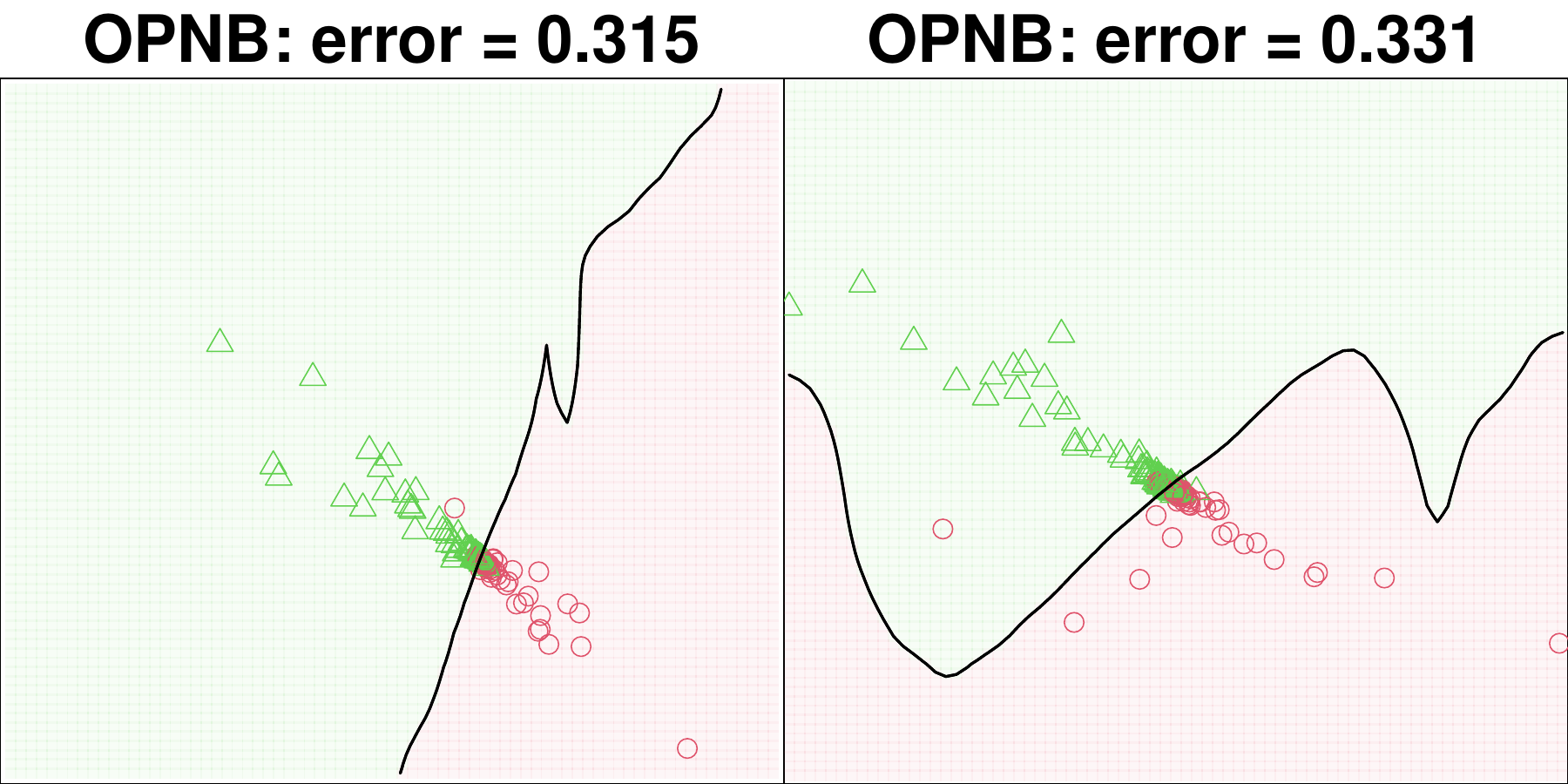}}
    \subfigure[Hill-Valley with Noise]{\includegraphics[width=0.48\linewidth]{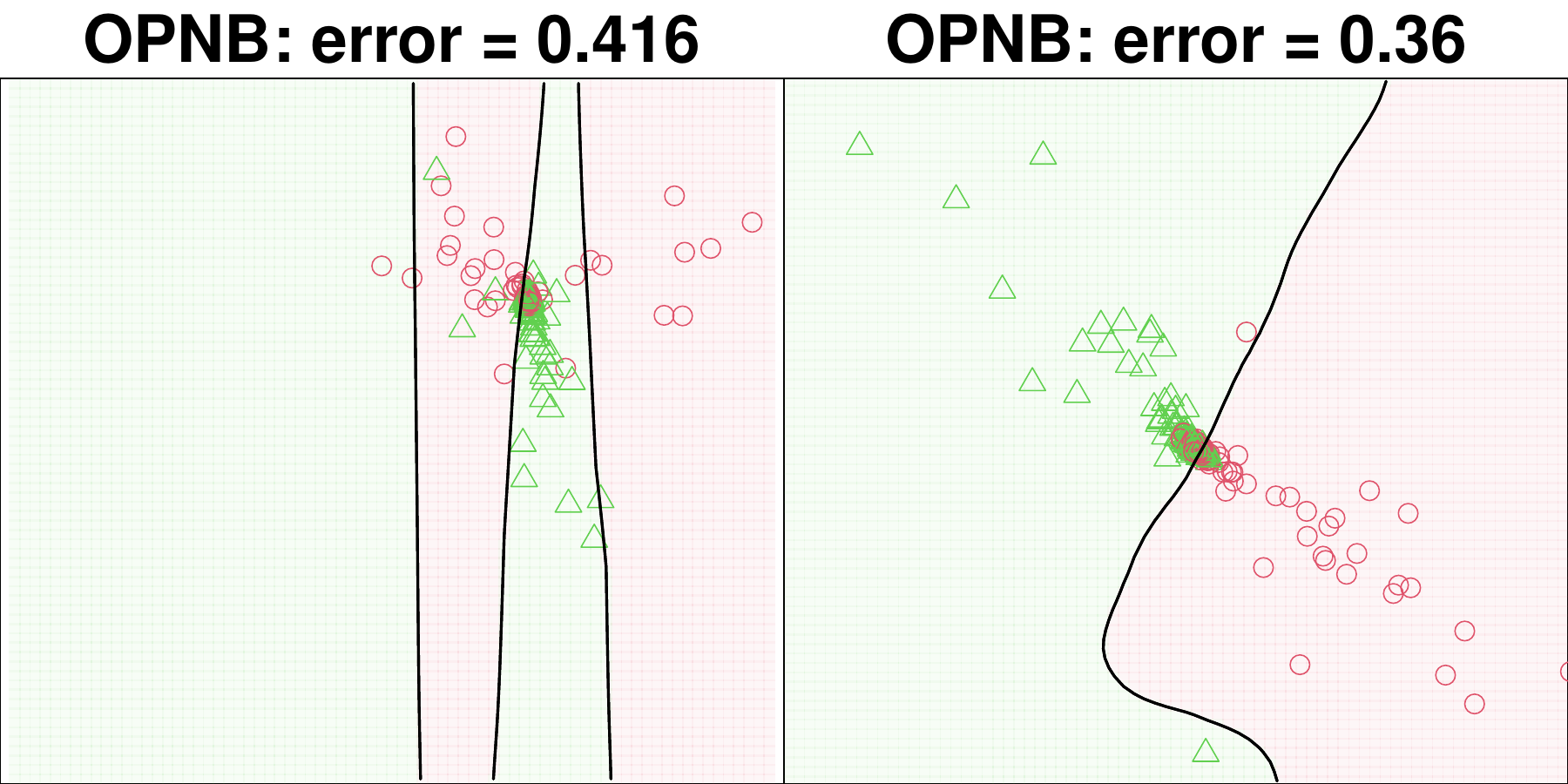}}
    \caption{OPNB projection plots for the Hill-Valley data sets. In each sub-figure the left plot shows the solution when $C$ is the total data covariance and the right plot when $C$ is the pooled within class covariance.}
    \label{fig:hillvalley_fixed}
\end{figure}

Fortunately for this particular problem the projection plots are typically clearly diagnostic, and there are simple strategies which are reasonably effective in mitigating this effect. We modified the objective function so that the penalty incorporates the scale of the projected data, and not only the magnitude of the projection matrix. Specifically, we replaced the objective with $\frac{1}{n} \ell(\V) - \lambda \mathrm{tr}\left(\V^\top C \V\right)$, where $C$ is either the (total) data covariance matrix or the pooled within class covariance matrix, and tr$(\cdot)$ is the trace operator. The solutions obtained with this simple modification are shown in Figure~\ref{fig:hillvalley_fixed}. For both settings of $C$ the solutions are vastly improved compared with the default implementation, and the performance is now more-or-less on par with the other models. It is worth pointing out that we did not do any selection of $\lambda$ for these experiments, and simply used $\lambda = 0.001$ which is our working default setting.

\subsubsection{Poor Initialisations}

As mentioned previously we used principal components analysis (PCA) to initialise the optimisation of the OPNB projections. This was done partly for reproducibility, and because investigations of the projection plots is clearly not possible for all 162 data sets and all 10 train/test splits of each.

\begin{figure}
    \centering
    \subfigure[PCA Initialisation]{\includegraphics[width=0.27\linewidth]{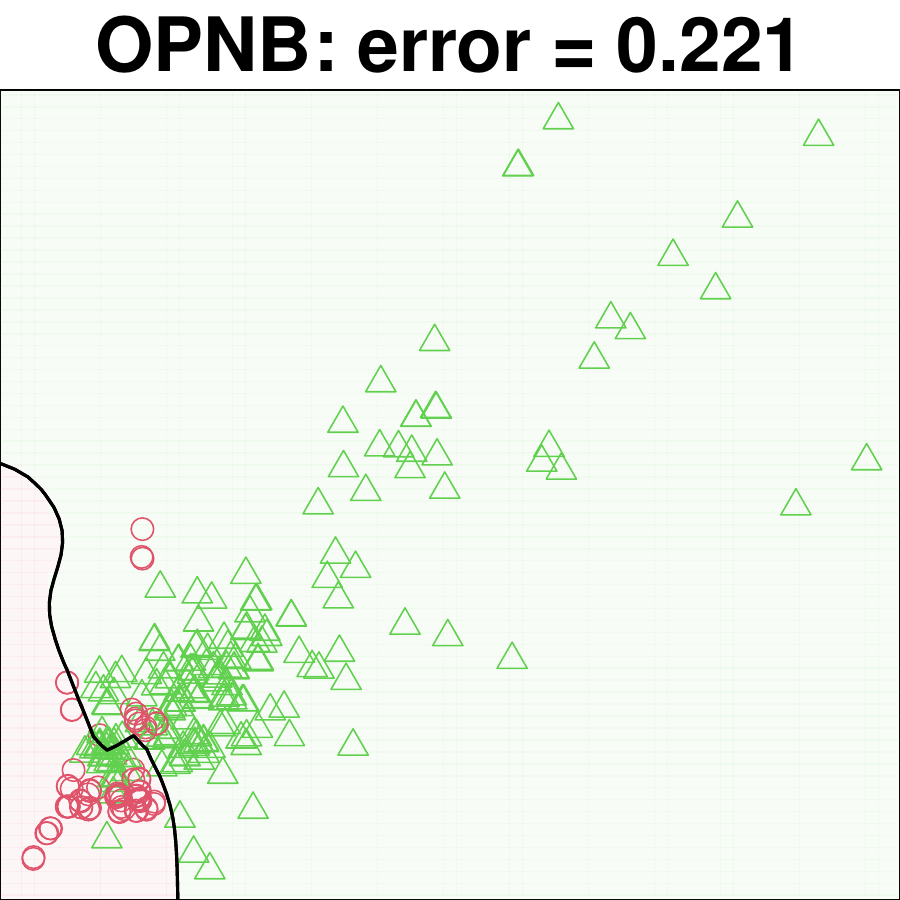}\label{fig:spectf1}}
    \subfigure[Random Initialisation]{\includegraphics[width=0.27\linewidth]{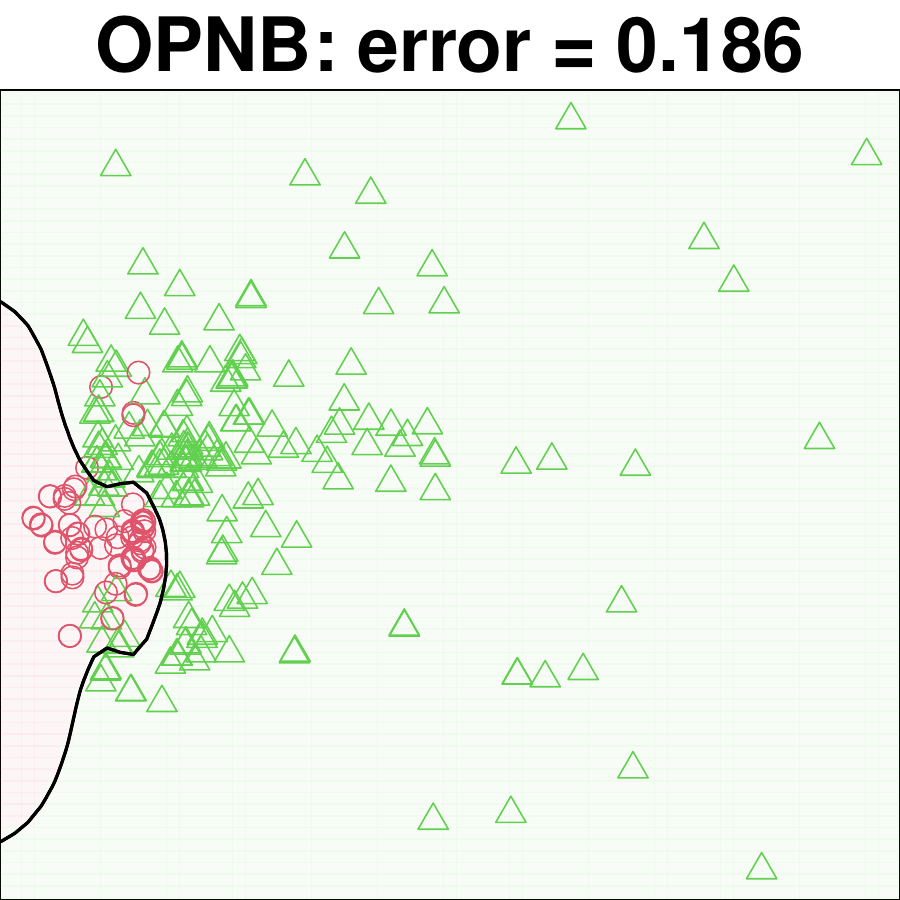}\label{fig:spectf2}}
    \caption{Two OPNB models obtained from the 44 dimensional spectf data set. (a) The default solution obtained from PCA initialisation; (b) the best, in terms of objective function value, from 20 random initialisations.}
    \label{fig:spectf}
\end{figure}

Here we briefly discuss an example where the initialisation with PCA led to poor performance. Whereas in the last example the problem was easily identified by the class decision boundaries, in this case the projections of the training observations themselves are indicative of a (potentially) poor solution. Figure~\ref{fig:spectf} shows plots associated with two OPNB models fit to the ``spectf'' data set\footnote{While we were not able to find meta data for this dataset in the PMLB git repo, it is likely related to the SPECTF Heart dataset from the UCI repository \url{https://doi.org/10.24432/C5N015}. Each patient in the dataset is labeled as either normal or abnormal (binary classification) with 44 features derived from an image taken from the patient. Data is available for 267 patients, less than the 349 contained in the PMLB dataset.}.

Recall that the \nb~model is most appropriate when the class conditional distributions approximately factorise along the cardinal basis directions, and the objective of OPNB is to obtain an alternative (optimal) basis for this factorisation. Although there are subtleties associated with the following statement which we will not go into, the unsuitability of a projection of the data in this regard may be assessed visually by whether the individual classes show any strong dependence. Figure~\ref{fig:spectf1} shows the solution obtained from the default initialisation with PCA, where the class shown with \textcolor{green}{$\triangle$} has an obvious diagonal orientation, indicating strong dependence between the projections along the first two OPNB projections. On the other hand, the solution shown in Figure~\ref{fig:spectf2} is the best (based on the value of the objective function at convergence) from 20 random initialisations. The class distributions are much more appealing from the point of view of the \nb~factorisation, and it is unsurprising that the test error (shown at the top of the plots) is superior. It is worth pointing out too that the objective function value at convergence for the random solution is indeed superior to that arising from the PCA initialisation.

\section{Discussion} \label{sec:conclusions}

The \nb~model has substantially reduced complexity when compared with other non-parametric discriminant models. However, inevitably the associated reduction in variance comes at the expense of, sometimes substantial, model bias. Unfortunately, although the bias/variance trade-off can be traversed by modifying the bandwidth parameter(s), the rigid model formulation means this traversal is highly limited. In this paper we introduced an intuitive way in which the flexibility of the \nb~model can be increased within an optimisation based framework, allowing better use of the model's effective degrees of freedom. By selecting an optimal basis for the factorisation of class conditional densities, this approach has been shown to achieve strong performance across a large collection of benchmark data sets. An investigation into the performance of the method, with reference to the characteristics of the data sets considered, suggests that our approach is flexible in its ability to model scenarios with complex decision boundaries, and is robust to the presence of large numbers of categorical covariates. Moreover, the visualisations offered by the model, due to its reliance on projection pursuit, helps to diagnose limitations in the model as well as aid interpretation of the model's decision boundaries.

\bibliographystyle{plainnat}

\end{document}